\documentclass[11pt,a4paper]{article}
\usepackage[hyperref]{acl2019}
\usepackage{times}
\usepackage{latexsym}
\usepackage{amsmath}
\usepackage{url}
\usepackage{color,soul}
\usepackage{todonotes}
\usepackage{multirow}
\usepackage{subcaption}
\usepackage{comment}
\usepackage[linesnumbered,ruled,vlined]{algorithm2e}

\aclfinalcopy 

\newcommand{\Note}[2]{} 
\newcommand{\SideNote}[2]{} 
\renewcommand{\Note}[2]{\todo[color=#1,size=\small, inline=true]{#2}} 
\renewcommand{\SideNote}[2]{\todo[color=#1,size=\small]{#2}} %

\title{Deep Structured Neural Network for Event Temporal Relation Extraction} 

\author{Rujun Han\thanks{\quad The authors contribute equally, alphabetical order.} , I-Hung Hsu\footnotemark[1] , Mu Yang\\ 
\texttt{\{rujunhan, ihunghsu, yangmu\} @isi.edu}\\
\textbf{Aram Galstyan, Ralph Weischedel, Nanyun Peng} \\
\texttt{\{galstyan, weisched, npeng\} @isi.edu}\\
Information Sciences Institute, University of Southern California}

\date{}

\begin{document}
\maketitle
\begin{abstract}
    We propose a novel deep structured learning framework for event temporal relation extraction. The model consists of 1) a recurrent neural network (RNN) to learn scoring functions for pair-wise relations, and 2) a structured support vector machine (SSVM) to make joint predictions. The neural network automatically learns representations that account for long-term contexts to provide robust features for the structured model, while the SSVM incorporates domain knowledge such as transitive closure of temporal relations as constraints to make better globally consistent decisions. By jointly training the two components, our model combines the benefits of both data-driven learning and knowledge exploitation. 
    Experimental results on three high-quality event temporal relation datasets (TCR, MATRES, and TB-Dense) demonstrate that incorporated with pre-trained contextualized embeddings, the proposed model achieves significantly better performances than the state-of-the-art methods on all three datasets. We also provide thorough ablation studies to investigate our model.
\end{abstract}

\begin{figure}[th!]
\centering
    \begin{subfigure}[b]{\columnwidth}
    \centering
    \includegraphics[width=\columnwidth]{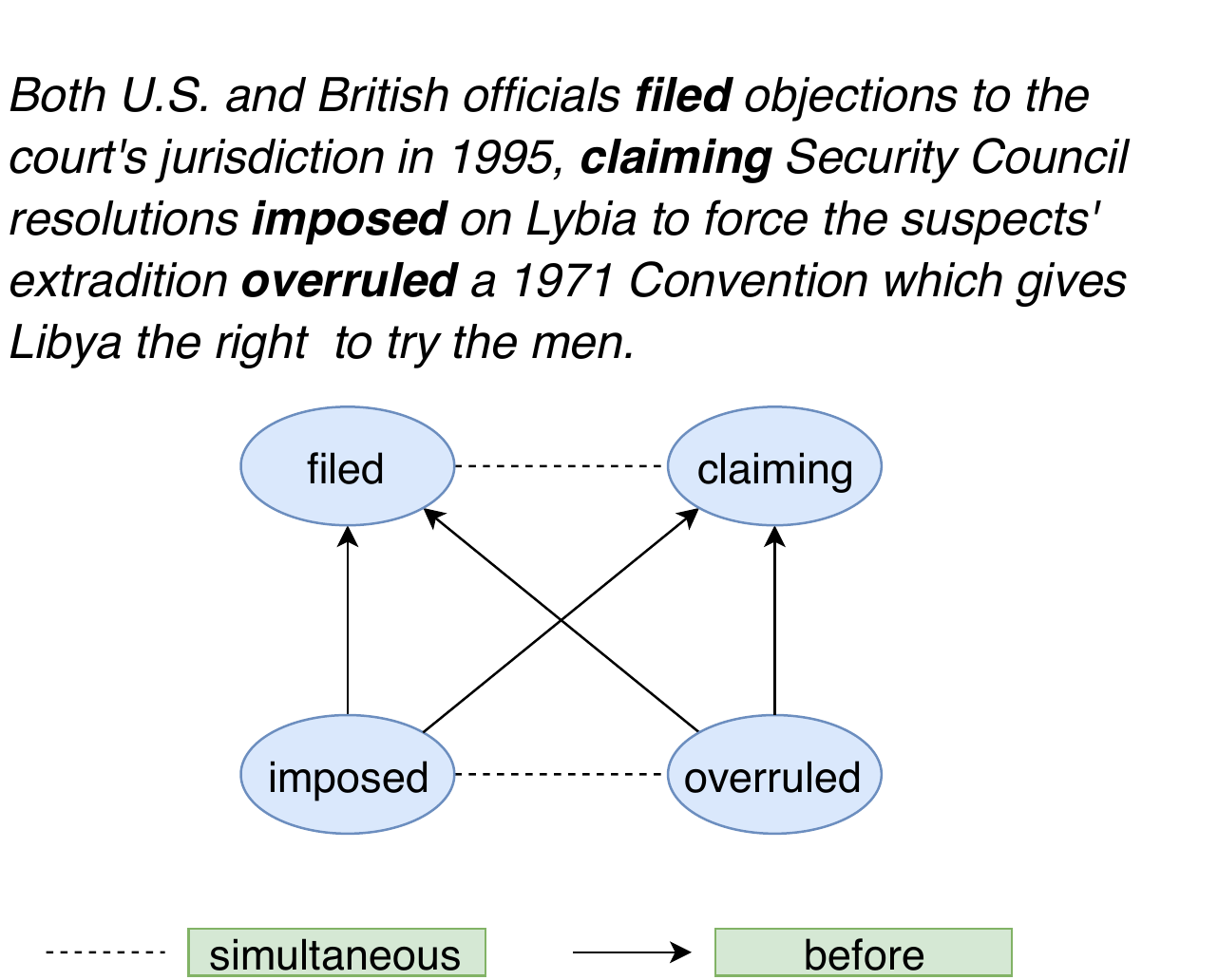}
    \caption{Ground-Truth Sub-graph}
    \label{fig:ex12}
    \end{subfigure}
    \begin{subfigure}[b]{0.9\columnwidth}
    \centering
    \includegraphics[width=0.9\columnwidth]{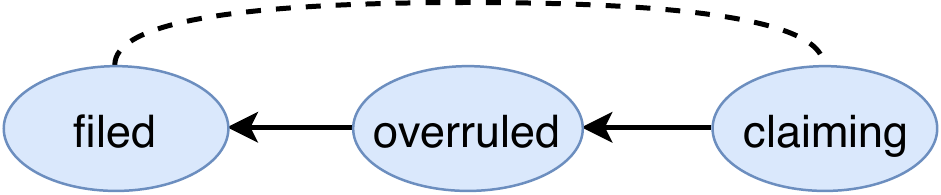}
    \caption{Local Model Predictions}
    \label{fig:ex13}
    \end{subfigure}
    \begin{subfigure}[b]{0.9\columnwidth}
    \centering
    \includegraphics[width=0.9\columnwidth]{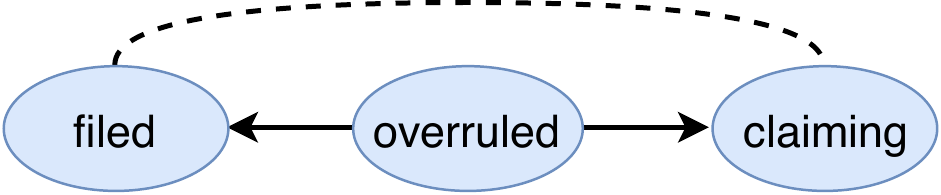}
    \caption{Structured Model Predictions}
    \label{fig:ex14}
    \end{subfigure}
\vspace{-0.1cm}
\caption{An illustration of a paragraph with its partial temporal graph. (a) shows the Ground-Truth temporal graph, in which case temporal relation \textit{SIMULTANEOUS} and \textit{BEFORE} are presented. (b) and (c) are the local and structured predictions, respectively, for three of the event pairs in (a). Local predictions are incompatible with temporal transitivity rule: \textbf{overruled} has to be \textit{BEFORE} \textbf{claiming} if \textbf{overruled} is \textit{BEFORE} \textbf{filed} and \textbf{filed} is \textit{SIMULTANEOUS} with \textbf{claiming}. The structured model achieves coherence by reversing the temporal order between \textbf{claiming} and \textbf{overruled}.}
\label{fig:ex1} 
\vspace{-.5em}
\end{figure} 

\section{Introduction}
Event temporal relation extraction aims at building a graph where nodes correspond to events within a given text, and edges reflect temporal relations between the events.
Figure~\ref{fig:ex12} illustrates an example of such graph for the text shown above. Different types of edges specify different temporal relations: the event \textbf{filed} is \textit{SIMULTANEOUS} with \textbf{claiming}, \textbf{overruled} is \textit{BEFORE} \textbf{claiming}, and \textbf{overruled} is also \textit{BEFORE} \textbf{filed}.
Temporal relation extraction is beneficial for many downstream tasks such as question answering, information retrieval, and natural language generation. 
An event graph can potentially be leveraged to help time-series forecasting and provide guidances for natural language generation. The CaTeRs dataset~\citep{MostafazadehX2016} which annotates temporal and causal relations is constructed for this purpose. 

A major challenge in temporal relation extraction stems from its nature of being a {\em structured prediction} problem. Although a relation graph can be decomposed into individual relations on each event pair, any {\em local} model that is not informed by the whole event graph  will usually fail to make globally consistent predictions, thus degrading the overall performance.  Figure~\ref{fig:ex13} gives an example where the \emph{local model} classifies the relation between \textbf{overruled}
 and \textbf{claiming} incorrectly as it only considers pairwise predictions: graph temporal transitivity constraint is violated given the relation between \textbf{filed} and \textbf{claiming} is \textit{SIMULTANEOUS}. In Figure~\ref{fig:ex14}, the \emph{structured model} changes the prediction of relation between \textbf{overruled} and \textbf{claiming} from \textit{AFTER} to \textit{BEFORE} to ensure compatibility of all predicted edge types.

Prior works on event temporal relation extraction mostly formulate it as a pairwise classification problem~\citep{S13-2002, laokulrat-EtAl:2013:SemEval-2013, chambers:2013:SemEval-2013, ChambersTBS2014} disregarding the global structures. \newcite{BramsenDLB2006, ChambersJ2008, DoLuRo12, NingWuRo18, P18-1212} explore leveraging global \emph{inference} to ensure consistency for all pairwise predictions. There are a few prior works that directly model global structure in the training process~\cite{yoshikawa2009jointly,NingFeRo17,leeuwenberg2017structured}. However, these structured models rely on hand-crafted features using linguistic rules and local-context, which cannot adequately capture potential long-term dependencies between events. In the example shown in Figure~\ref{fig:ex1}, \textbf{filed} occurs in much earlier context than \textbf{overruled}. Thus, incorporating long-term contextual information can be critical for correctly predicting temporal relations.

In this paper, we propose a novel deep structured learning model to address the shortcomings  of the previous methods. Specifically, we adapt the structured support vector machine (SSVM) \citep{Finley:2008:TSS:1390156.1390195} to incorporate linguistic constraints and domain knowledge for making joint predictions on events temporal relations. Furthermore, we augment this framework with recurrent neural networks (RNNs) to learn long-term contexts. Despite the recent success of employing neural network models for event temporal relation extraction~\cite{tourille2017neural, cheng2017classifying, meng2017temporal, meng2018context}, these systems make pairwise predictions, and do not take advantage of problem structures.

We develop a joint end-to-end {\em training} scheme that enables the feedback from global structure to directly guide neural networks to learn representations, and hence allows our deep structured model to combine the benefits of both data-driven learning and knowledge exploitation. In the ablation study, we further demonstrate the importance of each global constraints, the influence of linguistic features, as well as the usage of contextualized word representations in the local model.

To summarize, our main contributions are: 
\begin{itemize}
\vspace{-0.13cm}
\setlength\itemsep{0em}
    \item We propose a deep SSVM model for event temporal relation extraction. 
    \item We show strong empirical results and establish new state-of-the-art for three event relation benchmark datasets. 
    \item Extensive ablation studies and thorough error analysis are conducted to understand the capacity and limitations of the proposed model, which provide insights for future research on temporal relation extraction.

\end{itemize}

\section{Related Work}
\paragraph{Temporal Relation Data.} Temporal relation corpora such as TimeBank \citep{PustejovskyX2003} and RED \citep{O'Gorman2016} facilitate the research in temporal relation extraction. The common issue in these corpora is missing annotation. Collecting densely annotated temporal relation corpora with all event pairs fully annotated has been reported to be a challenging task as annotators could easily overlook some pairs \citep{P14-2082, Bethard:2007:TTI:1304608.1306306, ChambersTBS2014}. 
TB-Dense dataset mitigates this issue by forcing annotators to examine all pairs of events within the same or neighboring sentences. Recent data construction efforts such as MATRES~\citep{P18-1212} and TCR~\citep{NingWuRo18} further enhance the data quality by using a multi-axis annotation scheme and adopting start-point of events to improve inter-annotator agreements. However, densely annotated datasets are relatively small both in terms of number of documents and event pairs, which restricts the complexity of machine learning models used in previous research. 

\paragraph{Event Temporal Relation Extraction} 
The series of TempEval competitions \citep{Verhagen:2007:STT:1621474.1621488, Verhagen:2010:STT:1859664.1859674, S13-2001} attract many research interests in predicting event temporal relations. 
Early attempts \cite{Mani:2006:MLT:1220175.1220270, Verhagen:2007:STT:1621474.1621488, Chambers:2007:CTR:1557769.1557820,Verhagen:2008:TPT:1599288.1599300} only use local pair-wise classification with hand-engineered features. Later efforts, such as ClearTK \citep{S13-2002}, UTTime \citep{laokulrat-EtAl:2013:SemEval-2013}, and NavyTime \cite{chambers:2013:SemEval-2013} improve earlier work by feature engineering with linguistic and syntactic rules. A noteworthy work, CAEVO \citep{ChambersTBS2014}, builds a pipeline with ordered sieves. 
Each sieve is either a rule-based classifier or a machine learning model; sieves are sorted by precision, i.e. decisions from a lower precision classifier cannot contradict those from a higher precision model. 

More recently, neural network-based methods have been employed for event temporal relation extraction~\cite{tourille2017neural, cheng2017classifying, meng2017temporal,han2019contextualized} which achieved impressive results. However, they all treat the task as a pairwise classification problem. \newcite{meng2018context} considered incorporating global context for pairwise relation predictions, but they do not explicitly model the \emph{output} graph structure for event temporal relation.

There are a few prior works exploring structured learning for temporal relation extraction~\cite{yoshikawa2009jointly,NingFeRo17,leeuwenberg2017structured}. However, their local models use hand-engineered linguistic features. Despite the effectiveness of hand-crafted features in previous research, the design of features usually fails to capture long-term context in the discourse. Therefore, we propose to enhance the hand-crafted features with contextual representations learned through RNN models and develop an integrated joint training process.
\section{Methods}


We adapt the notations from \citet{P18-1212}, where $\mathcal{R}$ denotes the set of all possible relations; $\mathcal{E}$ denotes the set of all event entities. 

\subsection{Deep SSVM}
Our deep SSVM model adapts the SSVM loss as
{\small
\begin{align} \label{eq:loss}
\mathcal{L} = & \sum_{n = 1}^l \frac{1}{M^n} \big[\max \big(0, \Delta(\boldsymbol{y}^n, \hat{\boldsymbol{y}}^n) \text{ } + \\
& S(\hat{\boldsymbol{y}}^n; \boldsymbol{x}^n) - S(\boldsymbol{y}^n;\boldsymbol{x}^n) \big)\big] + C||\Phi||^2\nonumber,
\end{align}
}
where $\Phi$ denotes model parameters, $n$ indexes instances, $M^n$ is the number of event pairs in instance $n$. $\boldsymbol{y}^n,\hat{\boldsymbol{y}}^n$ denote the gold and predicted global assignments for instance $n$, each of which consists of $M^n$ one hot vectors $\boldsymbol{y}_{i,j}^n, \boldsymbol{\hat{y}}_{i,j}^n \in \{0, 1\}^{|\mathcal{R}|}$ representing true and predicted relation labels for event pair $i,j$ respectively. $\Delta(\boldsymbol{y^n}, \hat{\boldsymbol{y}}^n)$ is a distance measurement between the gold and the predicted assignments; we simply use hamming distance.  $C$ is a hyper-parameter to balance the loss and the regularizer, and $S(\boldsymbol{y}^n;\boldsymbol{x}^n)$ is a pair-wise scoring function to be learned. 

The intuition behind the SSVM loss is that it requires the score of gold output structure $\boldsymbol{y}^n$ to be greater than the score of the best output structure under the current model $\hat{\boldsymbol{y}}^n$ with a margin $\Delta(\boldsymbol{y}^n, \hat{\boldsymbol{y}}^n)$\footnote{Note that if the best prediction is the same as the gold structure, the margin is zero.}, or else there will be some loss. 

The major difference between our deep SSVM and the traditional SSVM model is the scoring function. Traditional SSVM uses a linear function over hand-crafted features to compute the scores, whereas we propose to use a RNN for estimation. 
\begin{figure}
    \centering
    \includegraphics[width=\columnwidth]{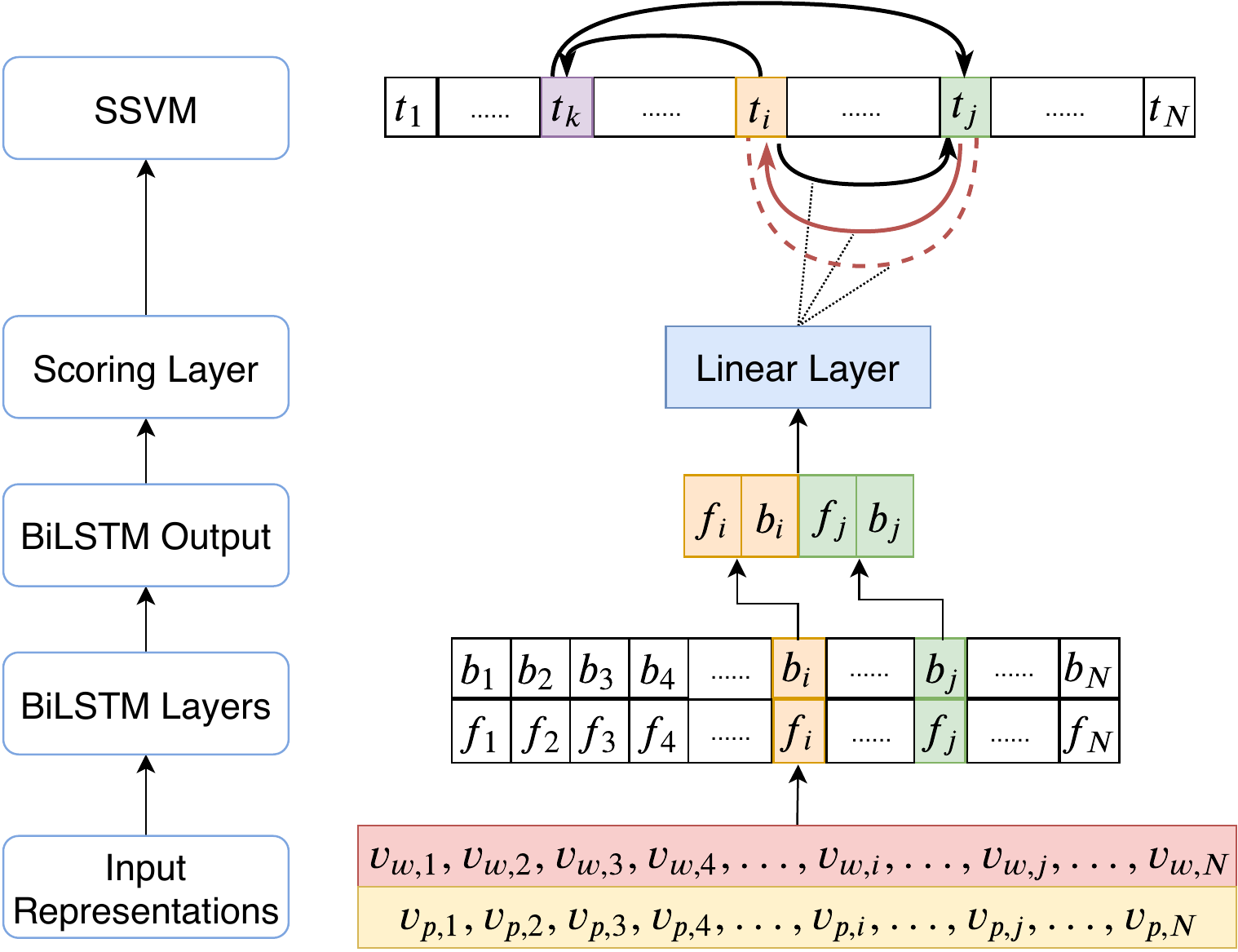}
    \caption{\label{fig:bisltm} An overview of the proposed deep structured event relation extraction framework. The input representations consist of BERT representations ($v_{w,k}$) and POS tag embeddings ($v_{p,k}$). They are concatenated to pass through BiLSTM layers and classification layers to get pairwise local scores. Incompatible local pairwise prediction (denoted by red lines) is corrected by the SSVM layer. Edge notation follows Figure \ref{fig:ex1} and $t_1, ... t_N$ denote tokens in the input sentence. }
\end{figure}
\subsection{RNN-Based Scoring Function} \label{sec:rnn_local}
We introduce a RNN-based pair-wise scoring function to learn features in a data-driven way and capture long-term context in the input. The local neural architecture is inspired by prior work in entity relation extraction such as ~\citet{S17-2098}.
As shown in Figure~\ref{fig:bisltm}, the input layer consists of word representations and part-of-speech (POS) tag embeddings of each token in the input sentence, denoted as $v_{w,k}$ and $v_{p,k}$ respectively.\footnote{Following the convention of event relation prediction literature \citep{ChambersTBS2014, P18-1212, NingWuRo18}, we only consider event pairs that occur in the same or neighboring sentences, but the architecture can be easily adapted to the case where inputs are longer than two sentences.} The word representations are obtained via pre-trained BERT \cite{devlin2018bert}\footnote{We use pre-trained bert-base-uncased model from \href{https://github.com/huggingface/pytorch-transformers}{https://github.com/huggingface/pytorch-transformers}.} model and are fixed throughout training, while the POS tag embeddings are tuned.
The word and POS tag embeddings are concatenated to represent an input token, and then fed into a Bi-LSTM layer to get contextualized representations.

We assume the events are labeled in the text and use indices $i$ and $j$ to denote the tokens associated with an event pair ($i, j$) $\in \mathcal{E}\mathcal{E}$ in the input sentences of length N. For each event pair ($i,j$), we take the forward and backward hidden vectors corresponding to each event, namely $f_i, b_i, f_j, b_j$ to encode the event tokens. 
These hidden vectors are then concatenated to form the input to the final linear layer to produce a softmax distribution over all possible pair-wise relations, which we refer to as RNN-based \textbf{scoring function}.



\subsection{Inference} \label{sec:inference}
The inference is needed both during training to obtain $\hat{\boldsymbol{y}}^n$ in the loss function (Equation~\ref{eq:loss}), as well as during the test time to get globally compatible assignments. 
The inference framework is established by constructing a global objective function using scores from local model and imposing several global constraints: symmetry and transitivity as in \citet{BramsenPYR12, ChambersJ2008, DenisM2008, DoLuRo12, NingFeRo17, han2019joint}, as well as linguistic rules and temporal-causal constraints proposed by \citet{P18-1212} to ensure global consistency. In this work, we incorporate the \textbf{symmetry}, \textbf{transitivity}, and \textbf{temporal-causal} constraints. 

\paragraph{Objective Function.} 
The objective function of the global inference maximizes the score of global assignments as specified in Equation~\ref{eq:infObj}\footnote{The objective function is specified on the instance level.}. 

{\small
\begin{align} \label{eq:infObj}
\hat{y} = & \arg \max \sum_{(i,j) \in \mathcal{E}\mathcal{E}} \sum_{r \in \mathcal{R}} y^r_{i,j} S(y^r_{i,j};\boldsymbol{x})
\end{align}
\[
\text{s.t.      } y^r_{i,j} \in \{0, 1\} \text{  , } \sum_{r \in \mathcal{R}} y^r_{i,j} = 1,
\]}
where $y^r_{i,j}$ is a binary indicator specifying if the global prediction is equal to a certain label $r \in \mathcal{R}$ and $S(y^r_{i,j},\boldsymbol{x}), \forall r \in \mathcal{R}$ is the scoring function obtained from the local model. The output of the global inference $\hat{y}$ is a collection of optimal label assignments for all event pairs in a fixed context. The constraint following immediately from the objective function is that the global inference should only assign one label to each pair of sample inputs. 

\paragraph{Symmetry and Transitivity constraint.}
The symmetry and transitivity constraints are used across all models and experiments in the paper. They can be specified as follows:

{\small
\[
\forall (i,j), (j,k) \in \mathcal{E}\mathcal{E}, y^r_{i,j} = y^{\bar{r}}_{j,i}, \text{ (symmetry)}
\]
\[
y^{r_1}_{i,j} + y^{r_2}_{j,k} - \sum_{r_3 \in Trans(r_1, r_2)} y^{r_3}_{i,k} \leq 1. \text{ (transitivity)}
\]
}

Intuitively, the symmetry constraint forces two pairs with opposite order to have reversed relations. For example, if $r_{i,j}$ = \textit{BEFORE}, then $r_{j,i}$ = \textit{AFTER}. Transitivity constraint rules that if ($i,j$), ($j,k$) and ($i,k$) pairs exist in the graph, the label (relation) prediction of ($i,k$) pair has to fall into the transitivity set specifying by ($i,j$) and ($j,k$) pairs. The full transitivity table can be found in \citet{P18-1212}.

\paragraph{Temporal-causal Constraint.}
The temporal-causal constraint is used for the TCR dataset which is the only dataset in our experiments that contains causal pairs and it can written as:
\[
y^c_{i,j} = y^{\bar{c}}_{j,i} \leq y^b_{i,j}, \text{ } \forall (i,j) \in \mathcal{E}\mathcal{E},
\]
where c and $\bar{c}$ correspond to the label \textit{CAUSES} and \textit{CAUSED\_BY}, and b represents the label \textit{BEFORE}. This constraint specifies that if event i causes event j, then i has to occur before j. Note that this constraint only has 91.9\% accuracy in TCR data \citep{P18-1212}, but it can help improve model performance based on our experiments. 
\subsection{Learning}
We develop a two-state learning approach to optimize the neural SSVM. We first train the local scoring function without feedback from global constraints. In other words, the local neural network model is optimized using only pair-wise relations in the first stage by minimizing cross-entropy loss. During the second stage, we switch to the global objective function in Equation~\ref{eq:loss} and re-optimize the network to adjust for global properties\footnote{We experiment with optimizing SSVM loss directly, but model performance degrades significantly. We leave further investigation for future work.}. We denote the local scoring model in the first stage as local model, and the final model as global model in the following sections.

\section{Experimental Setup}
In this section, we describe the three datasets that are used in the paper. Then we define the evaluation metrics. Finally, we provide details regarding our model implementation and experiments.

\subsection{Data}\label{data_intro}
Experiments are conducted on TB-Dense, MATRES and TCR datasets and an overview of data statistics are shown in Table~\ref{tab:data}. We focus on event relation, thus, all numbers refer to $\mathcal{E}\mathcal{E}$ pairs\footnote{For TCR, we also include causal pairs in the table.}. Note that in all three datasets, event pairs are always annotated by their appearance order in text, i.e. given a labeled event pair ($i$, $j$), event $i$ always appears prior to event $j$ in the text. Following \citet{meng2017temporal}, we augment event pairs with flipped-order pairs. That is, if a pair ($i$, $j$) exists, pair ($j$, $i$) is also added to our dataset with the opposite label. The augmentation is applied to training and development split, but test set remains unaugmented\footnote{It is noted that if symmetric constraint is applied, scores for testing on augmented or unaugmented set are equal.}.

 \begin{table}[!]
    \smallskip
    \small
 	\centering
 	\begin{tabular}{l|c|c|c} \hline
 	& TB-Dense & MATRES & TCR \\ \hline\hline
    \multicolumn{4}{c}{\textbf{\# of Documents}} \\ \hline
 	Train & 22 &  22 & 20\\
 	Dev  &  5 &  5 & 0\\
    Test & 9 &  9 & 5 \\\hline\hline
    \multicolumn{4}{c}{\textbf{\# of Pairs}} \\ \hline
 	Train & 4032 &  1098 & 1992 \\
 	Dev & 629 & 229  & 0 \\
 	Test & 1427 & 310  & 1008 \\\hline
 	\end{tabular}
   	\caption{Data Overview}
   	\label{tab:data}
 \end{table}
 
\paragraph{TB-Dense} \citep{P14-2082} is based on TimeBank Corpus but addresses the sparse-annotation issue in the original data by introducing the \textit{VAGUE} label and requiring annotators to label all pairs of events/times in a given window.

\paragraph{MATRES} \citep{NingWuRo18} is based on TB-Dense data, but filters out non-verbal events. The authors project events on multiple axes and only keep those in the main-axis. These two factors explain the large decrease of event pairs in Table~\ref{tab:data}. Start-point temporal scheme is adopted when out-sourcing the annotation task, which contributes to the performance improvement of machine learning models built on this dataset . 

\paragraph{TCR} \citep{P18-1212} follows the same annotation scheme for temporal pairs in MATRES. It is also annotated with causal pairs. To get causal pairs, the authors select candidates based on EventCausality dataset \citep{do2011minimally}. 

\subsection{Evaluation Metrics}
To be consistent with the evaluation metrics used in baseline models, we adopt two slightly different calculations of metrics.
\paragraph{Micro-average} For all datasets, we compute the micro-average scores. For densely annotated data, the micro-average metric should share the same precision, recall and F1 scores. However, since VAGUE pairs are excluded in the micro-average calculations of TCR and MATRES for fair comparisons with the baseline models, the micro-average for precision, recall and F1 scores are different when reporting results for the two datasets.
\paragraph{Temporal Awareness (TE3)} For TB-Dense dataset, TE3 evaluation scheme \citep{S13-2001} is also adopted in previous research \citep{NingFeRo17, P18-1212}. TE3 score not only takes into account of the number of correct pairs but also capture how ``useful'' a temporal graph is. We report this score for TB-Dense results only. For more details of this metric, please refer to the original paper~\citep{S13-2001}.

\begin{table}
    \small
 	\centering
 	\begin{tabular}{l|c|c|c} \hline
 	& TB-Dense & MATRES & TCR \\ \hline\hline
    \multicolumn{4}{c}{\textbf{Local Model}} \\ \hline
 	hid\_size & 60 &  40 & 30\\
 	dropout  &  0.5 &  0.7 & 0.5\\
 	BiLSTM layers & 1 & 2 & 1\\
 	learning rate & 0.002 &  0.002 & 0.002 \\\hline\hline
    \multicolumn{4}{c}{\textbf{Structured Learning}} \\ \hline
 	learning rate & 0.05 &  0.08 & 0.08 \\
 	decay  & 0.7 & 0.7  & 0.9 \\\hline
 	\end{tabular}
   	\caption{Best hyper-parameters}
   	\label{tab:hyper}
\end{table}
\subsection{Implementation Details} \label{sec:impl} 
Since our work focuses on event relations, we build our models to predict relations between $\mathcal{E}\mathcal{E}$ pairs only when conducting experiments. Thus, all micro-average F1 scores only consider $\mathcal{E}\mathcal{E}$ pairs. Note that there are also time entities labeled in the TB-Dense denoted as $\mathcal{T}$. $\mathcal{E}\mathcal{T}$ and $\mathcal{T}\mathcal{T}$ pairs are generally easier to predict using rule-based classifiers or date normalization technique \citep{DoLuRo12, ChambersTBS2014, MirzaT2016}. To be consistent with the baseline models \cite{P18-1212, NingWuRo18} for TB-Dense data, we add $\mathcal{E}\mathcal{T}$ and $\mathcal{T}\mathcal{T}$ pairs for TE3 evaluation metric\footnote{We rely on annotated data to distinguish different pair types, i.e. $\mathcal{E}\mathcal{E}$, $\mathcal{E}\mathcal{T}$ and $\mathcal{T}\mathcal{T}$ are assumed to be given.}.

In the two-stage learning procedure, the local model is trained by minimizing cross-entropy loss with Adam optimizer.
We use pre-trained BERT embedding with 768 dimensions as the input word representations and one-layer MLP as the classification layer.
As for the structured learning stage, we observe performance boost by switching from Adam optimizer to SGD optimizer with decay and momentum\footnote{The weight decay in SGD is exactly the value $C$ in Equation~\ref{eq:loss}. We set the momentum in SGD as 0.9 in all datasets.}. To solve ILP in the inference process specified in Section~\ref{sec:inference}, we leverage off-the-shelf solver provided by Gurobi optimizer, i.e. the best solutions from the Gurobi optimizer are inputs to the global training. 

The hyper-parameters are chosen by the performance on the development set\footnote{We randomly select 4 documents from the training set as development set for TCR.}, and the best combination of hyper-parameters can be found in Table~\ref{tab:hyper}. We run experiments on 3 different random seeds and report the average results.

Note that for TCR data, we need a separate classifier for causal relations. Because of small amount of causal pairs, we simply build an independent final linear layer apart from the original linear layer in Figure~\ref{fig:bisltm}. In other words, there are two final linear layers: only one of them is active when training temporal or causal pairs.

\begin{figure}[!]
\centering
\includegraphics[width=\columnwidth]{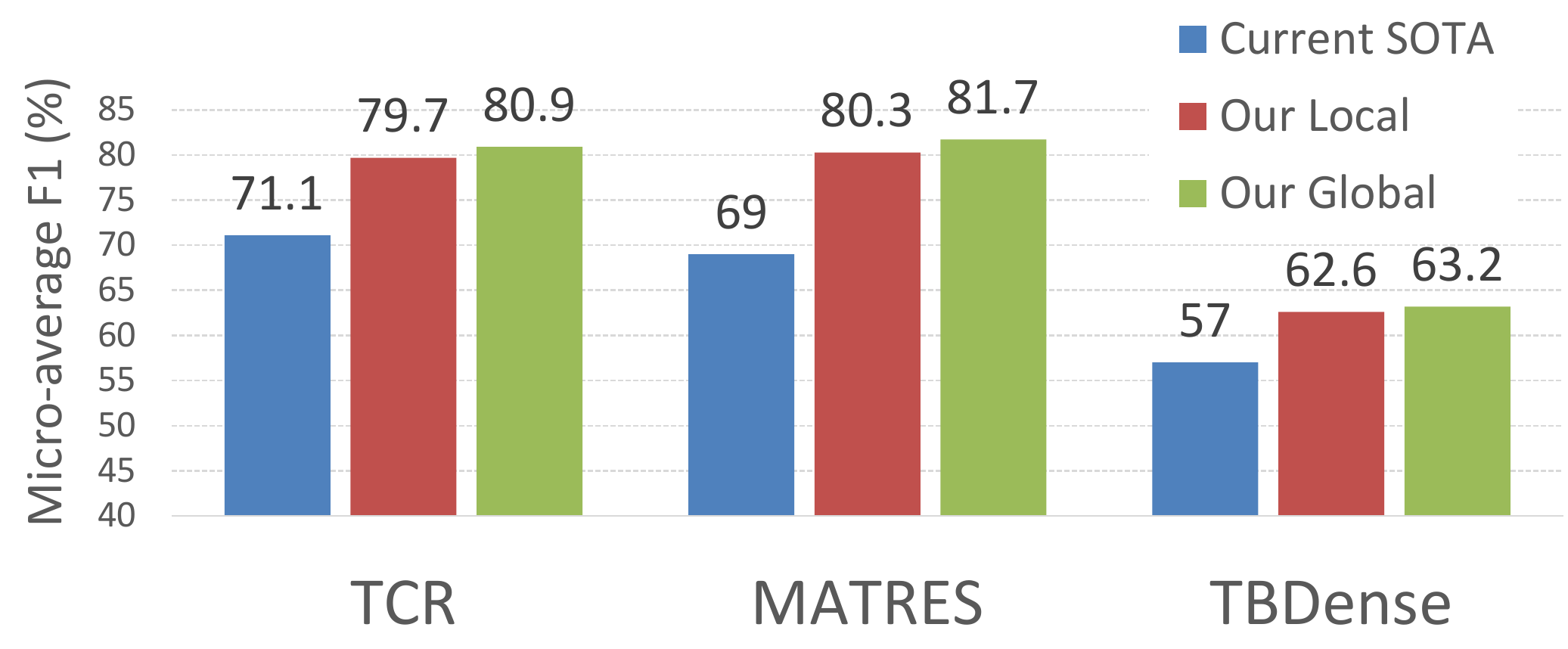}
\caption{
\label{fig:overall-results}
Model Performance (F1 Score) Overview. Our local and global models' performances are averaged over 3 different random seeds for robustness.}
\end{figure} 
\section{Results and Analysis}
Figure~\ref{fig:overall-results} shows an overview of our model performance on three different datasets. As the chart illustrates, our RNN-based local models outperform state-of-the-art (SOTA) results and the global models further improve the performance over local models across all three datasets.
\subsection{TCR}
Detailed model performances for the TCR dataset are shown in Table~\ref{tab:tcr-breakdown}.  We only report model performance on temporal pairs. Both of our local and global models outperform the baseline. Our global model is able to improve overall model performance by more than 1.2\% over our local model; per McNemar's test, this improvement is statistically significant (with p-value $< 0.01$). 
\begin{table}[!]
 	\centering
 	\footnotesize
 	\setlength{\tabcolsep}{0.5em}
 	\begin{tabular}{|l|c|c|c|c|c|c|} \hline
 	&\multicolumn{3}{|c|}{\textbf{Local Model}} & \multicolumn{3}{|c|}{\textbf{Global Model}}\\ \cline{2-7}
 	&\textbf{P}& \textbf{R} & \textbf{F1} & \textbf{P}& \textbf{R} & \textbf{F1}\\ \hline
 	Before & 82.1 & 86.9 & 84.3 & 81.3 & 90.0 & 85.4\\
 	After & 67.1 & 73.2 & 69.7 & 70.9 & 70.9 & 70.9\\
 	Simultaneous & 0.0 & 0.0 & 0.0 & 0.0 & 0.0 & 0.0\\
 	Vague & 0.0 & 0.0 & 0.0 & 0.0 & 0.0 & 0.0\\\hline
 	\textbf{Micro-average} & 77.1 & 82.5 & 79.7 & 78.2 & 83.9 & \textbf{80.9**}\\ \hline\hline
 	\multicolumn{6}{|l}{\textbf{\citet{P18-1212}}} & 71.1\\ \hline
 	\end{tabular}
   	\caption{Model Performance Breakdown for TCR. To make fair comparison, we exclude VAGUE pairs in Micro-average score, which is why P, R and F1 are different. **indicates global model outperforms local model with p-value $< 0.01$ per McNemar's test.}
   	\label{tab:tcr-breakdown}
\end{table}

\subsection{MATRES}
Detailed model performances for the MATRES dataset performances can be found in Table~\ref{tab:mat-breakdown}. 
Similar to TCR, both our local and structured models outperform this baseline and the global model is able to improve overall model performance by 1.4\%; per McNemar's test, this improvement is statistically significant (with p-value $< 0.05$).
\begin{table}[!]
 	\centering
 	\footnotesize
 	\setlength{\tabcolsep}{0.5em}
 	\begin{tabular}{|l|c|c|c|c|c|c|} \hline
 	&\multicolumn{3}{|c|}{\textbf{Local Model}} & \multicolumn{3}{|c|}{\textbf{Global Model}}\\ \cline{2-7}
 	&\textbf{P}& \textbf{R} & \textbf{F1} &\textbf{P}& \textbf{R} & \textbf{F1}\\ \hline
 	Before & 79.7 & 88.1 & 83.6 & 80.1 & 89.6 & 84.6\\
 	After & 70.5 & 83.3 & 76.3 & 72.3 & 84.8 & 78.0\\
 	Simultaneous & 0.0 & 0.0 & 0.0 & 0.0 & 0.0 & 0.0\\
 	Vague & 0.0 & 0.0 & 0.0 & 0.0 & 0.0 & 0.0\\\hline
 	\textbf{Micro-average} & 76.2 & 84.9 & 80.3 & 77.4 & 86.4 & \textbf{81.7*}\\ \hline\hline
 	\multicolumn{6}{|l} {\textbf{\citet{NingWuRo18}}} & 69\\ \hline
 	\end{tabular}
   	\caption{Model Performance Breakdown for MATRES. Again, we exclude VAGUE pairs in Micro-average score. * indicates global model outperforms local model with p-value $< 0.05$ per McNemar's test.}
   	\label{tab:mat-breakdown}
\end{table}
\subsection{TB-Dense}
Table~\ref{tab:tbd-breakdown} shows the breakdown performance for all labels as well as the improvement from local to global model by adopting the two-stage structured learning method in TB-Dense dataset. Both our local and global models are able to outperform previous SOTA in micro-average metric (reported by \citet{meng2018context}) or in TE3 metric (results from \citet{P18-1212}).

Per McNemar's test, the improvements from local to global model only has p-value $=0.126$, so we are not able to conclude that the improvement is statistically significant. We think one of the reasons is the large share of VAGUE pairs (42.6\%). VAGUE pairs make our transitivity rules less conclusive. For example, if $\mathcal{R}(e1, e2)$ = BEFORE and $\mathcal{R}(e2, e3)$ = VAGUE, $\mathcal{R}(e1, e3)$ can be any relation types. Moreover, this impact is magnified by our local model's prediction bias towards VAGUE pairs. As we can see in Table~\ref{tab:tbd-breakdown}, the recall score for VAGUE pairs are much higher than other relation types, whereas precision score is moderate. Our global model leverages local output structure to enforce global prediction consistency, but when local predictions contain many VAGUE pairs, it introduces lots of noise too. 

To make fair comparison between our model and the best reported TE3 F1 score from \citet{P18-1212}, we follow their strategy and add CAEVO system's predictions on $\mathcal{T}\mathcal{T}$ and $\mathcal{E}\mathcal{T}$ pairs in the evaluation. The scores are shown in Table~\ref{tab:tbd-breakdown}. Our overall system outperforms the baseline over 10\% for both micro-average and TE3 F1 scores.
\begin{table}[t!]
 	\centering
 	\footnotesize
 	\setlength{\tabcolsep}{0.5em}
 	\begin{tabular}{|l|r|r|r|r|r|r|} \hline
 	&\multicolumn{3}{|c|}{\textbf{Local Model}} & \multicolumn{3}{|c|}{\textbf{Global Model}}\\ \cline{2-7}
 	 & P & R & F1 & P & R & F1\\ \hline
 	Before & 73.5 & 52.7 & 61.3 & 71.1 & 58.9 & 64.4\\
 	After & 71.6 & 60.8 & 65.3 & 75.0 & 55.6 & 63.5\\
 	Includes & 17.5 & 4.8 & 7.4 & 24.6 & 4.2 & 6.9\\
 	Is\_Include & 69.1 & 4.4 & 8.0 & 57.9 & 5.7 & 10.2\\
 	Simultaneous & 0.0 & 0.0 & 0.0 & 0.0 & 0.0 & 0.0\\
 	Vague & 57.9 & 81.5 & 67.7 & 58.3 & 81.2 & 67.8\\\hline
 	\textbf{Micro-average} &  \multicolumn{3}{|r}{62.6} & \multicolumn{3}{|r|}{\textbf{63.2}} \\\hline
 	\multicolumn{6}{|l}{\textbf{\citet{ChambersTBS2014}}} & 49.4  \\
 	\multicolumn{6}{|l}{\textbf{\citet{cheng2017classifying}}} & 52.9  \\ 
 	\multicolumn{6}{|l}{\textbf{\citet{meng2018context}}} & \textbf{57.0}  \\ \hline \hline
 	\multicolumn{7}{|c|}{\textbf{TE3 Metrics}} \\ \hline
 	$\mathcal{E}\mathcal{E}$ only & 62.1 & 61.9 & 62.2 & 62.7 & 58.9 & \textbf{62.5} \\
 	$+$ $\mathcal{E}\mathcal{T}, \mathcal{T}\mathcal{T}$ & 58.6 & 63.6 & 61.0 & 59.0 & 64.0 & 61.4 \\\hline
 	\multicolumn{6}{|l}{\textbf{\citet{P18-1212}}} & 52.1  \\\hline
 	\end{tabular}
   	\caption{Model Performance Breakdown for TB-Dense (all values are percentage). Upper Table: for event pairs only, we adopt standard Micro-average score. Lower Table: TE3 refers to the temporal awareness score adopted by TE-3 Workshop. To make fair comparison with \citet{P18-1212}, we add CEAVO predictions on $\mathcal{E}\mathcal{T}$ and $\mathcal{T}\mathcal{T}$ pairs back into the calculation.}
   	\label{tab:tbd-breakdown}
\end{table}

\subsection{Error Analysis}
To understand why both the local and structured models make mistakes, we randomly sample 50 pairs from 345 cases where both models' predictions are incorrect among all 3 random seeds. We analyze these pairs qualitatively and categorize them into four cases as shown in Table~\ref{tab:tbd_err_category}, with each case (except other) paired with an example.

The first case illustrates that correct prediction requires broader contextual knowledge. For example, the gold label for \textbf{transition} and \textbf{discuss} is \textit{BEFORE}, where the nominal event \textbf{transition} refers to a specific era in history that ends before \textbf{discuss} in the second sentence. Human annotators can easily infer this relation based on their knowledge in history, but it is difficult for machines without prior knowledge. We observe this as a very common mistake especially for pairs with nominal events. As for the second case shows that negation can completely change the temporal order. The gold label for the event pair \textbf{planned} and \textbf{plans} is \textit{AFTER} because the negation token \textbf{no} postpones the event \textbf{planned} indefinitely. Our models do not pick up this signal and hence predict the relation as \textit{VAGUE}. 

Finally, ``intention'' events could make temporal relation prediction difficult \citep{NingWuRo18}. Case 3 demonstrates that our models could ignore the ``intention'' tokens such as \textbf{aimed at} in the example and hence make an incorrect prediction \textit{VAGUE} between \textbf{doubling} and \textbf{signed}, whereas the true label is \textit{AFTER} because \textbf{doubling} is an intention that has not occurred.

 \begin{table}[!]
 \setlength{\textfloatsep}{5pt}
    \smallskip
    \small
 	\centering
 	\begin{tabular}{l} \hline
 	\textbf{Case 1 (32\%):} Connection with broader context \\ \hline
    The program also calls for coordination of economic \\
    reforms and joint improvement of social programs in the \\
    two countries, where many people have become \\
    impoverished during the chaotic post - Soviet \textbf{transition} \\
    to capitalism. Kuchma also planned to visit Russian gas \\
    giant Gazprom, most likely to \textbf{discuss} Ukraine's DLRS \\
    1.2 billion debt to the company. \\ \hline\hline
    \textbf{Case 2 (20\%):} Negation \\ \hline
    Annan has \textbf{no} trip \textbf{planned} so far. Meanwhile, Secretary \\
    of State Madeleine Albright, Berger and Defense \\
    Secretary William Cohen announced \textbf{plans} to travel \\
    to an unnamed city in the us heartland next week, \\
    to explain to the American people just why military \\
    force will be necessary if diplomacy fails. \\ \hline\hline
    \textbf{Case 3 (14\%):} Intention Axis \\ \hline
    A major goal of Kuchma's four - day state visit was the \\
    signing of a 10-year economic program \textbf{aimed at} \\ \textbf{doubling} the two nations' trade turnover, which fell to \\
    DLRS 14 billion last year, down DLRS 2.5 billion from \\
    1996. The two presidents on Friday \textbf{signed} the plan, \\
    which calls for cooperation in the metallurgy, fuel, \\
    energy, aircraft building, missile, space and chemical \\
    industries. \\ \hline\hline
    \textbf{Case 4: (34\%)} Other \\ \hline
 	\end{tabular}
   	\caption{Error Categories and Examples in TB-Dense}
   	\label{tab:tbd_err_category}
 \end{table}

\section{Ablation Studies} \label{sec:abl_study}
Although we have presented strong empirical results, the isolated contribution of each component of our model has not been investigated. In this section, we perform  a though ablation study to understand the importance of  structured constraints, linguistic features, and the BERT representations. 

\subsection{Effect of the structured constraints}
One of our core claims is that our learning benefits from modeling the structural constraints of event temporal graph. To study the contribution of structured constraints, we provide an ablation study on two constraints that are applied to all three datasets: Symmetry and Transitivity.

A straightforward ablation study on symmetric constraint is to remove it from our global inference step. However, even though we eliminate symmetric constraint explicitly in global inference, it is utilized implicitly in our data augmentation steps (Section ~\ref{data_intro}). To better understand the benefits of the symmetry constraints, we study both the contribution of \textit{explicitly} applying symmetry constraint in our SSVM as well as its \textit{implicit} impact in data augmentation.

Hence, in this section, we view a pair with original order and flipped order as different instances for learning and evaluation. We denote the pairs with original order as ``\textbf{forward}" data, their flipped-order counterparts as ``\textbf{backward}" data, and their combinations as ``\textbf{both-way}" data.

We train four additional models to study the impacts of symmetry and transitivity constraints: 1) local model trained on forward data; 2) global model with transitivity constraint trained on forward data; 3) local model trained on both-way data; 4) global model with transitivity constraint trained on both-way data, denoted as $M_1$, $M_2$, $M_3$, $M_4$ respectively. $M_1$ and $M_2$ are models that do not apply any symmetric property; $M_3$ and $M_4$ are models that utilize symmetric property implicitly. 

Additionally, evaluation setup should be re-scrutinized if we remove the symmetry constraints. In the standard evaluation setup of prior works, evaluation is only performed on the pairs with their original order (forward data) in text. This evaluation assumes a model will work equally well for both forward and backward data, which certainly holds when we explicitly impose symmetry constraints. However, as we can observe in the later analysis, this assumption fails when we remove symmetry constraints.
To demonstrate the improvement of model robustness over backward data, we propose to test the model on both forward and both-way data. If a model is robust, it should perform well on both scenarios. 
\begin{table}[!]
    \resizebox{\columnwidth}{!}{
    \def\arraystretch{1.3}
    \begin{tabular}{ l | c | c | c | c | c | c } \hline
    & \multicolumn{2}{c|}{TB-Dense} & \multicolumn{2}{c|}{Matres} & \multicolumn{2}{c}{TCR} \\ \cline{2-7}
    &$\overrightarrow{Test}$ &$\overleftrightarrow{Test}$ &$\overrightarrow{Test}$ &$\overleftrightarrow{Test}$ &$\overrightarrow{Test}$ &$\overleftrightarrow{Test}$ \\ \hline
    $M_1:\ \overrightarrow{L}$ & 62.9 & 61.9 & 80.4 & 74.7 & 80.5 & 75.7\\
    $M_2:\ \overrightarrow{L}$ + $T$  & \textbf{63.2} &62.0 & \textbf{81.7} &  75.7  & \textbf{81.0} & 76.3\\
    $M_3:\ \overleftrightarrow{L}$  & 62.6 &62.7 & 80.3 &80.4 & 79.7 &79.6\\
    $M_4:\ \overleftrightarrow{L}$ + $T$  & 63.1 & 63.0& 81.4 &81.4 & 80.3 &80.2\\
    \hline
    \multirow{2}{0.12\textwidth}{$\overleftrightarrow{L}$ + $S$ + $T$ (Proposed)}
    &\multirow{2}{*}{\textbf{63.2}} &\multirow{2}{*}{\textbf{63.2}} &\multirow{2}{*}{\textbf{81.7}} &\multirow{2}{*}{\textbf{81.7}} &\multirow{2}{*}{80.9} &\multirow{2}{*}{\textbf{80.9}} \\
    & & & & & & \\
    \hline
    \end{tabular}}
    \caption{Ablation over global constraints: Symmetry and Transitivity. Test is conducted on forward test set and both-way test set, which are denoted as $\overrightarrow{Test}$ and $\overleftrightarrow{Test}$ respectively. The local models trained on forward data and both-way data are denoted as $\overrightarrow{L}$ and $\overleftrightarrow{L}$. Symmetry and Transitivity constraints are denoted as $S$ and $T$. The results demonstrate that symmetry and transitivity constraints both improve model's performance.}
    \label{tab:abl_constraints}
\end{table}

We summarize our analysis of the results in Table \ref{tab:abl_constraints} (F1 scores) as follows:
\begin{itemize}
\vspace{-0.13cm}
\setlength\itemsep{0em}
    \item Impact of Transitivity: By comparing $M_1$ with $M_2$ and $M_3$ with $M_4$, the consistent improvements across all three datasets demonstrate the effectiveness of global transitivity constraints.
    \item Impact of Implicit Symmetry (data augmentation): Examining the contrast between $M_1$ and $M_3$ as well as $M_2$ and $M_4$, we can see significant improvements in both-way evaluation despite slight performance drops in forward evaluation. These comparisons imply that data augmentation can help improve model robustness. Note that \newcite{meng2018context} leveraged this data augmentation trick in their model.
    \item Impact of Explicit Symmetry: By comparing the proposed model with $M_4$, the consistent improvements across all datasets demonstrate the benefit of using explicit symmetry property.
    \item Model Robustness: Although $M_1$ and $M_2$ show competitive results when evaluated on forward test data, their performance degrade significantly in the both-way evaluation. In contrast, the proposed model achieves strong performances in both test scenarios (best F1 scores except for one), and hence proves the robustness of our proposed method.

\end{itemize}

\subsection{Effect of linguistic features} \label{sec:abl_feat}
Previous research establish the success of leveraging linguistic features in event relation prediction. One advantage of leveraging contextualized word embedding is to provide rich semantic representation and could potentially avoid the usage of extra linguistic features. Here, we study the impact of incorporating linguistic features to our model by using simple features provided in the original datasets: token distance, tense and polarity of event entities. These features are concatenated with the Bi-LSTM hidden states before the linear layer (i.e. $f_i$, $b_i$, $f_j$, $b_j$ in Figure \ref{fig:bisltm}). Table \ref{tab:abl_feat} shows the F1 scores of our local and global model using or not using linguistic features respectively. These additional features likely cause over-fitting and hence do not improve model performance across all three datasets we test. This set of experiments show that linguistic features do not improve the predicting power of our current framework. 

\begin{table}
    \centering
    \small
    \resizebox{\columnwidth}{!}{
    \begin{tabular}{ c c c c c c } 
    \hline
            &  local-   & global-  & local-    & global- \\
            & w/ feat. & w/ feat. & w/o feat. & w/o feat. \\
    \hline
    TB-Dense & 62.5    & 63.0    & 62.6     & \textbf{63.2} \\
    MATRES  &  81.4    & \textbf{81.7}    & 80.3      & \textbf{81.7} \\
    TCR     & 79.5    & 80.7    & 79.7     & \textbf{80.9} \\
    \hline
    \end{tabular}}
    \caption{Ablation study on linguistic feature usage. Additional linguistic features do not lead to significant improvement and even hurt performance in 2 out of 3 datasets. The results show that our proposed framework is semantic-rich and capable of avoiding the usage of additional linguistic feature.}
    \label{tab:abl_feat}
\end{table}

\subsection{Effect of BERT representations}
In this section, we explore the impact of contextualized BERT representations under our deep SSVM framework. We replace BERT representations with the GloVe~\cite{Pennington14glove:global} word embeddings. Table \ref{tab:abl_bert} shows the F1 scores of our local model and global model using BERT and GloVe\footnote{For GloVe model, additional linguistic features are used.} respectively. BERT improves the performance with a significant margin. Besides, even without BERT representations, our RNN-based local model and the deep structured global model still outperform (MATRES and TCR) or are comparable with (TB-Dense) current SOTA. These results confirm the improvements of our method.

\begin{table}
    \centering
    \footnotesize
    \resizebox{\columnwidth}{!}{
    \begin{tabular}{ c c c c c c } 
    \hline
            & previous & local- & global- & local- & global- \\
            & SOTA     & Glove  & Glove   & BERT   & BERT \\
    \hline
    TB-Dense &57.0    & 56.6   & 57.0    & 62.6  & \textbf{63.2} \\
    MATRES  & 69.0    & 71.8   & 75.6    & 80.3   & \textbf{81.7} \\
    TCR     & 71.1    & 73.5   & 76.5    & 79.7  & \textbf{80.9} \\
    \hline
    \end{tabular}}
    \caption{Ablation over word representation: BERT vs GloVe. Although BERT representation largely contributes to the performance boost, our proposed framework remains strong and outperforms current SOTA approaches when GloVe is used.}
    \label{tab:abl_bert}
\end{table}

\section{Conclusion}
In this paper, we propose a novel deep structured model based on SSVM that combines the benefits of structured models' ability to encode structure knowledge, and data-driven deep neural architectures' ability to learn long-range features. Our experimental results exhibit the effectiveness of this approach for event temporal relation extraction. 

One interesting future direction is further leveraging commonsense knowledge, domain knowledge in temporal relation, and linguistics information to create more robust and comprehensive global constraints for structured learning. 
Another direction is to improve feature representations by designing novel neural architectures that better capture negation and hypothetical phrases as discussed in error analysis. We plan to leverage large amount of unannotated corpora to help event temporal relation extraction as well.

\section*{Acknowledgments}
This work is partially funded by DARPA Contracts W911NF-15-1-0543 and an NIH R01 (LM012592).
The authors thank the anonymous reviewers for their helpful suggestions and members from USC PLUS lab for early feedbacks. 
Any opinions, findings, conclusions, or recommendations expressed here are those of the authors and do not necessarily reflect the view of the sponsors.
\bibliography{acl2019}
\bibliographystyle{acl_natbib}

\end{document}